\documentclass[conference]{IEEEtran}
\IEEEoverridecommandlockouts

\usepackage{cite}
\usepackage{amsmath,amssymb}
\usepackage{graphicx}
\usepackage{booktabs}
\usepackage{url}
\usepackage[hidelinks]{hyperref}

\graphicspath{{figures/}{./}}

\newcommand{\Jsum}{J_{\Sigma}}
\newcommand{\Jmax}{J_{\max}}
\newcommand{\Bset}{\mathcal{B}}
\newcommand{\paperfigure}[2]{%
  \IfFileExists{#1}{%
    \includegraphics[width=\columnwidth]{#1}%
  }{%
    \fbox{\parbox[c][#2][c]{0.94\columnwidth}{%
      \centering\footnotesize MATLAB figure pending\\[2pt]
      \texttt{\detokenize{#1}}}}%
  }%
}

\title{Bundle Length Tradeoffs in Decentralized Multi-Robot Task Allocation Under Degraded Communications}

\author{
\IEEEauthorblockN{James Lott\\ \small Student Member, IEEE}
\IEEEauthorblockA{\textit{Shiley Marcos School of Engineering}\\
\textit{University of San Diego}\\
San Diego, United States of America\\
jlott@sandiego.edu}
\and
\IEEEauthorblockN{Vahraz Honary\\ \small Senior Member, IEEE}
\IEEEauthorblockA{\textit{Shiley Marcos School of Engineering}\\
\textit{University of San Diego}\\
San Diego, United States of America\\
vhonary@sandiego.edu}
}

\begin{document}
\maketitle

\begin{abstract}
Bundle length $B$ is commonly fixed when configuring multi-task multi-robot task allocation (MRTA) algorithms. MinSum and MinMax are known to favor different task distributions, but the role of $B$ in this objective tradeoff has not been systematically characterized. Additionally, degraded-communication evaluations also often retain settings selected under ideal communication, leaving whether nominal bundle-length tuning transfers under message loss unresolved. We examine both questions for ACBBA, PI, and HIPC across six bundle lengths in 300 paired ten-target Collaborative Visit scenarios under ideal communication and 25\% Bernoulli packet loss. Under ideal communication, increasing $B$ from 1 to 12 reduces MinSum cost by 19.0\%, 23.0\%, and 31.8\% for ACBBA, PI, and HIPC, respectively, while increasing MinMax cost by 45.6\%, 94.3\%, and 67.6\%. Under packet loss, the lowest-mean MinSum setting shifts from $B=12$ to $B=2$ for ACBBA and PI. Repeated paired cross-fitting shows that retaining the ideal-network setting incurs held-out MinSum penalties of 14.4\% and 7.2\%, respectively, and increases MinMax cost by 30.0\% and 41.8\% relative to the loss-conditioned MinSum setting. HIPC retains a deep MinSum operating region, while the MinMax setting remains stable for all three allocators. Experiments at two additional target loads reproduce the ACBBA and PI MinSum shifts. 
\end{abstract}


\section{Introduction}

Decentralized multi-robot task allocation (MRTA) distributes allocation decisions among robots operating from local information and exchanged messages rather than a continuously available central scheduler \cite{Gerkey2004,Quinton2023}. In multi-task routing, an allocation update may assign each robot an ordered sequence of future tasks. $B$ is a common per-robot constraint defined as the maximum number of ordered future tasks retained in that robot's active bundle. Each allocator preserves its native rule for constructing the bundle while operating within the same bound, so $B$ controls the depth of the active multi-task route. Bundle depth can therefore affect both route construction and workload distribution. This connects bundle configuration to two standard routing objectives. MinSum minimizes aggregate team route cost, whereas MinMax minimizes the largest individual route cost and therefore emphasizes completion of the final robot \cite{PI,Otte2020}. Prior routing results show that spatial consolidation favors MinSum while degrading MinMax by increasing the longest route \cite{Patil2022}. Bundle-length configuration may induce the same distinction within a decentralized allocator by determining how much multi-task route structure is available at each allocation update.

MRTA studies under communication degradation introduce a separate configuration issue. Both simulation benchmarks and hardware-based studies have shown that communication degradation can materially affect decentralized allocation performance \cite{Nayak2020,Cao2025,mypaper2}. These studies commonly establish algorithm parameters under nominal communication and retain them while varying the network. Nayak et al. select iteration and bundle parameters using perfect-communication experiments and reuse those settings across subsequent communication models \cite{Nayak2020}. Bapat et al. hold iteration counts and bundle bounds constant across communication levels \cite{Bapat2022}, while recent comparisons similarly vary packet loss, delay, fading, or communication range under fixed algorithm configurations \cite{Cao2025,Verma2025}. This controlled design isolates the degradation of a specified nominal treatment. It does not, however, determine whether the same configuration remains performance-preferred after the communication condition changes.

Existing communication-aware studies establish that network quality can alter relative allocator performance \cite{Otte2020} and can motivate online selection among allocation methods \cite{Carrillo2021}. Separate work has varied bundle length to characterize its effects on communication and computation \cite{Sao2025}. These lines of research do not establish whether communication degradation changes the bundle-length response within an allocator or quantify the performance cost of transferring a nominally selected bundle length to a degraded network. To the best of our knowledge, no prior decentralized MRTA study has evaluated this configuration-transfer question using separate $B$ selection and evaluation scenarios.

We examine asynchronous consensus-based bundle allocation (ACBBA)\cite{ACBBA}, performance impact (PI)\cite{PI}, and hybrid information and plan consensus (HIPC) \cite{HIPC}. We apply the same per-robot bundle-length bound to all three. ACBBA populates its ordered bundle using marginal bids, PI uses marginal performance significance, and HIPC constructs a locally predicted team allocation whose per-robot task sequences are bounded by $B$ before each robot retains its own ordered bundle for plan consensus. $B$ is therefore a common per-robot retained task bound while the mechanism used to populate those tasks remains allocator-specific. Collaborative Visit provides a controlled known-target routing problem in which the allocators must partition multiple simultaneous visit obligations. It therefore exposes both aggregate MinSum efficiency and MinMax completion balance while allowing the same bundle-length constraint to be varied across the three methods.

The study addresses two sequential questions. First, how does bundle length shape MinSum and MinMax performance under ideal communication? Second, does the resulting bundle-length response transfer under packet loss? The primary experiment evaluates six bundle lengths on 300 paired ten-target scenarios under ideal delivery and 25\% independent Bernoulli loss. Repeated paired cross-fitting separates configuration selection from evaluation, workload-concentration measures examine the allocation mechanism, and exploratory five- and twenty-target experiments assess whether the communication-conditioned operating regions persist across target loads. The contributions are:

\begin{itemize}
    \item A paired 300-scenario characterization of MinSum and MinMax response across six bundle lengths for ACBBA, PI, and HIPC
    \item Evidence that bundle growth exchanges aggregate route efficiency for parallel completion, supported by workload concentration and held-out objective-transfer analysis
    \item Evidence that 25\% packet loss moves the MinSum-preferred bundle region for ACBBA and PI, including held-out transfer costs and a scoped robustness check at two additional target loads.
\end{itemize}

\section{Formulation and Experimental Design}

\subsection{Objectives}

Let $\mathcal{R}=\{1,\ldots,n\}$ be the robot team and let $L_i$ denote the grid steps taken by robot $i$ before mission completion. We evaluate
\begin{equation}
    \Jsum=\sum_{i\in\mathcal{R}}L_i,
    \qquad
    \Jmax=\max_{i\in\mathcal{R}}L_i .
    \label{eq:objectives}
\end{equation}
$\Jsum$ is the MinSum route cost, referred to as \textit{total effort} below. $\Jmax$ is the MinMax route cost, referred to as \textit{makespan}. 

\subsection{Bundle Length}

We test
\begin{equation}
    B\in\Bset=\{1,2,3,5,8,12\},
    \label{eq:bset}
\end{equation}
Let $P_i$ denote robot $i$'s active ordered future-task list. For every allocator, the experiment imposes the same constraint $|P_i|\leq B$. ACBBA and PI construct $P_i$ through their native marginal-bid and marginal-significance procedures. HIPC constructs a local team allocation with each predicted robot sequence capped at $B$, then retains robot $i$'s ordered sequence as its personal bundle. Thus, $B$ is the same controlled per-robot bundle-depth constraint in all three treatments; allocator-specific logic determines which tasks occupy the available positions. When fewer than $B$ uncompleted tasks remain, the bound is nonbinding, so $B=12$ is the effectively unrestricted endpoint in the five- and ten-target experiments. Table~\ref{tab:bundle} summarizes the native bundle construction under this common bound.

\begin{table}[t]
\caption{Application of the Common Bundle-Length Bound}
\label{tab:bundle}
\centering
\footnotesize
\setlength{\tabcolsep}{3.2pt}
\begin{tabular}{@{}p{0.17\columnwidth}p{0.76\columnwidth}@{}}
\toprule
Method & Native bundle construction under common bound $B$ \\
\midrule
ACBBA & Greedy marginal-insertion bundle and corresponding execution path \cite{ACBBA}. \\
PI & Ordered task list updated by marginal significance and consensus \cite{PI}. \\
HIPC & Per-robot sequence in the locally predicted team allocation is capped at $B$; the robot retains its own ordered sequence \cite{HIPC}. \\
\bottomrule
\end{tabular}
\end{table}

\subsection{Mission Design}

Four homogeneous robots operate on an obstacle-free $19\times19$ grid with deterministic, evenly spaced starts along the west boundary. Ten target cells are known at initialization, and the Collaborative Visit mission \cite{Nayak2020} terminates when all targets have been visited. Experiments use an asynchronous discrete-event simulator in which all allocators share the same four-neighbor A* motion planner, event scheduler, allocation triggers, completion logic, and metric collection. ACBBA asynchronously reconciles marginal-insertion bids, PI reconciles task significance, and HIPC uses a local team-level allocation before plan consensus. Ideal communication denotes lossless delivery while in the impaired condition, each message is independently dropped per receiver with probability $p_d=0.25$. All non-$B$ allocator and simulator parameters are held fixed across bundle lengths and communication conditions.

\subsection{Experimental Design}

The primary experiment uses a fully crossed, scenario-paired design. Each of 300 ten-target mission instances is evaluated for all three allocators, six bundle lengths, and two communication conditions. Reusing each mission instance across all configurations controls for scenario difficulty and permits paired comparisons of bundle-length and communication effects. The primary campaign therefore contains $3\times6\times2\times300=10{,}800$ completed system-level runs.

Two independent campaigns repeat the same design with 5 and 20 targets using 25 paired scenarios per load. These experiments contribute $3\times6\times2\times25=900$ runs at each load and are used only to assess whether the operating regions observed in the primary experiment persist as task load changes. They are not pooled with the ten-target data for configuration selection or held-out evaluation. The complete study comprises 12,600 runs.

\subsection{Response and Transfer Analysis}

Let $\bar{Y}_{a,o,c}(B)$ be the mean outcome for allocator $a$, objective $o\in\{\Sigma,\max\}$, and communication condition $c$. Figure~\ref{fig:response} reports percentage change from the $B=1$ mean within each allocator, objective, and network:
\begin{equation}
    Q_{a,o,c}(B)=100\left[
       \frac{\bar{Y}_{a,o,c}(B)}{\bar{Y}_{a,o,c}(1)}-1
    \right].
    \label{eq:response}
\end{equation}
Numerical comparisons use raw means.

Lowest-mean tested settings are supporting reference points, not universal optima. Let $\widehat B_{a,o,c}$ minimize the observed mean over $\Bset$, resolving exact ties toward smaller $B$. Objective transfer evaluates MinMax at the MinSum reference. Communication transfer evaluates the ideal-network reference under loss relative to the loss-conditioned reference:
\begin{equation}
 R_{a,o}=100\left[
 \frac{\bar{Y}_{a,o,\mathrm{loss}}(\widehat B_{a,o,\mathrm{ideal}})}
      {\bar{Y}_{a,o,\mathrm{loss}}(\widehat B_{a,o,\mathrm{loss}})}-1
 \right].
 \label{eq:transfer}
\end{equation}

All full-sample uncertainty intervals use 20,000 paired bootstrap resamples of scenario identifiers. For Fig.~\ref{fig:response}, each resample preserves the pairing between the tested $B$ and its $B=1$ reference and recomputes Eq.~\eqref{eq:response} from the resampled means; $Q_{a,o,c}(1)=0$ by construction. Transfer contrasts use paired-bootstrap ratio-of-means intervals. To separate selection from evaluation \cite{Eggensperger2019}, we also run 50 repetitions of paired ten-fold cross-fitting. Within each fold, the other 270 scenario identifiers select ideal- and loss-conditioned references; the held-out 30 loss scenarios evaluate Eq.~\eqref{eq:transfer}. The same fold structure is used for objective transfer under ideal communication: training folds select the MinSum- and MinMax-preferred bundle lengths, and the held-out fold evaluates the resulting MinMax penalty. All bundle lengths and both communication outcomes for a scenario remain in the same fold. We report 500 foldwise selections and aggregate each repetition over its 300 out-of-fold predictions.

Two checks support interpretation. First, objective transfer is repeated after excluding $B=1$. Second, workload concentration is measured by the Gini coefficient over tasks completed and movement contributed by the four robots. The 5- and 20-target extensions report only whether the lowest-mean reference changes.

\section{Results}

\subsection{Objective-Specific Bundle Depth}

\begin{figure}[t]
    \centering
    \paperfigure{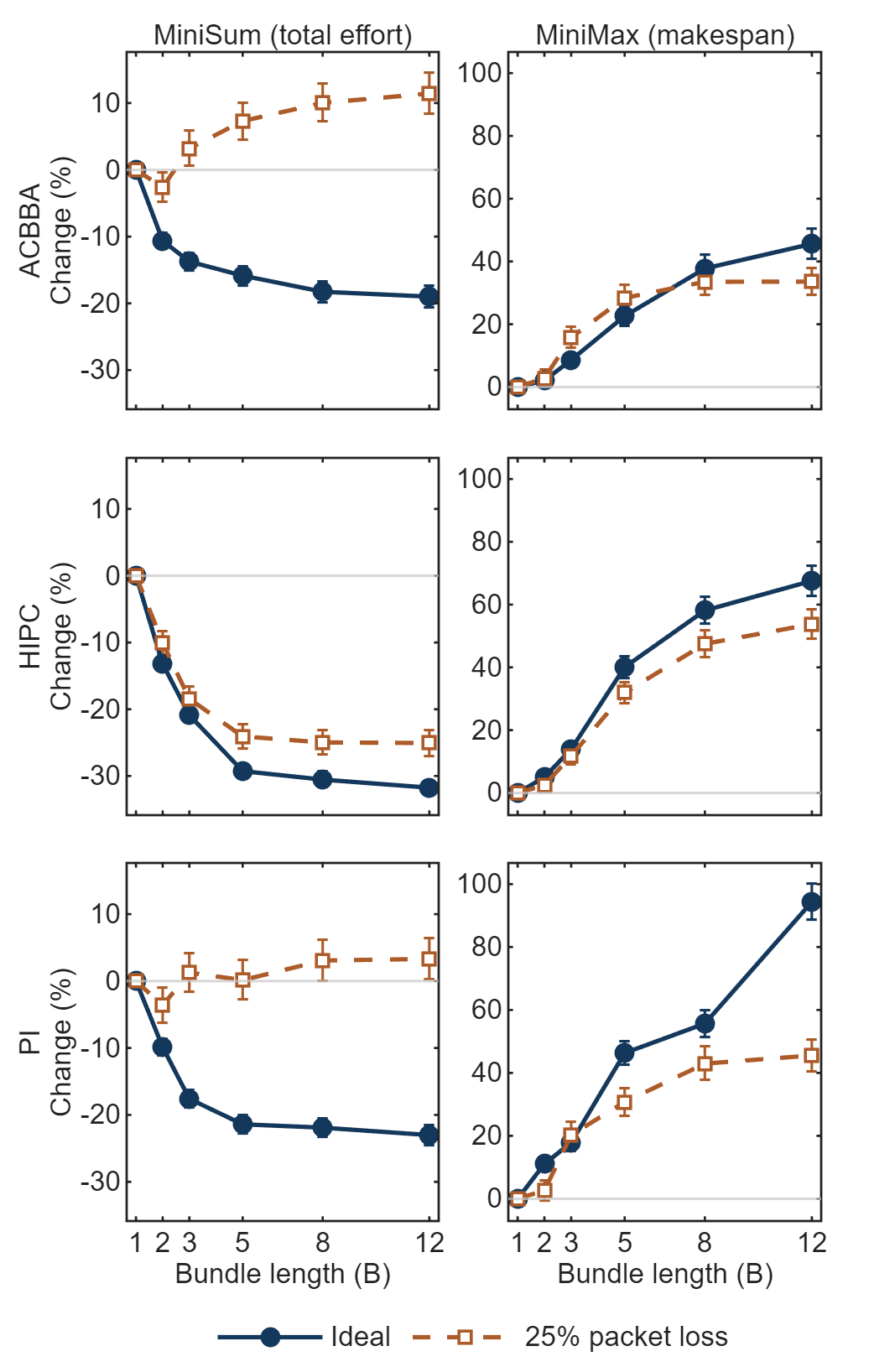}{4.75in}
    \caption{Ten-target response across bundle length. Rows are allocators; columns are MinSum and MinMax objectives. Values are percentage changes from $B=1$ within each communication condition. Solid curves denote ideal delivery and dashed curves denote 25\% packet loss. Error bars are 95\% paired-bootstrap intervals for the percentage change relative to $B=1$; the $B=1$ value is fixed at zero by construction.}
    \label{fig:response}
\end{figure}

Under ideal communication, all three allocators become more efficient in aggregate and less parallel as bundle length increases (Fig.~\ref{fig:response}). From $B=1$ to $B=12$, ACBBA total effort decreases from 80.93 to 65.56 steps ($-19.0\%$), while makespan increases from 23.41 to 34.09 steps ($+45.6\%$). HIPC effort decreases by 31.8\% while makespan increases by 67.6\%. PI effort decreases by 23.0\% while makespan increases by 94.3\%. The response curves show that deeper bundles consistently trade MinSum improvement for MinMax degradation; however, specific responses vary by algorithm.

The lowest ideal-network means occur at opposite ends of the tested range: $B=12$ for MinSum and $B=1$ for MinMax for all three allocators. Evaluating the MinSum reference for MinMax raises makespan by 45.6\% for ACBBA (95\% CI 40.9--50.4\%), 67.6\% for HIPC (62.7--72.4\%), and 94.3\% for PI (88.6--100.1\%). Repeated cross-fitting reproduces held-out penalties of 45.6\%, 67.6\%, and 94.4\%. After excluding $B=1$ as a single-task boundary, moving MinMax references to $B=2$, the resulting full-sample penalties remain large at 42.7\%, 59.6\%, and 75.0\%. The tradeoff is therefore not only a single-task boundary effect.

Additionally, from $B=1$ to $B=12$, target-workload Gini rises from 0.23 to 0.39 for ACBBA, 0.23 to 0.56 for HIPC, and 0.21 to 0.58 for PI. Movement-workload Gini reflects the same trend, rising from 0.07 to 0.33, 0.08 to 0.55, and 0.07 to 0.54, respectively. Deeper bundles reduce total movement while assigning a larger share of tasks and travel to fewer robots, a pattern consistent with the MinSum--MinMax separation.

\subsection{Effects of Packet Loss}

Introduction of packet loss changes the shape of the ACBBA and PI MinSum responses. Relative to ideal communication at the same $B$, ACBBA's effort degradation grows from 19.7\% at $B=1$ to 64.7\% at $B=12$; PI's grows from 32.7\% to 78.1\%. Their ideal MinSum curves descend through $B=12$, whereas their loss curves reach their lowest means at $B=2$ and then rise. HIPC's MinSum curve continues to favor the deep region under loss.

\begin{figure}[t]
    \centering
    \paperfigure{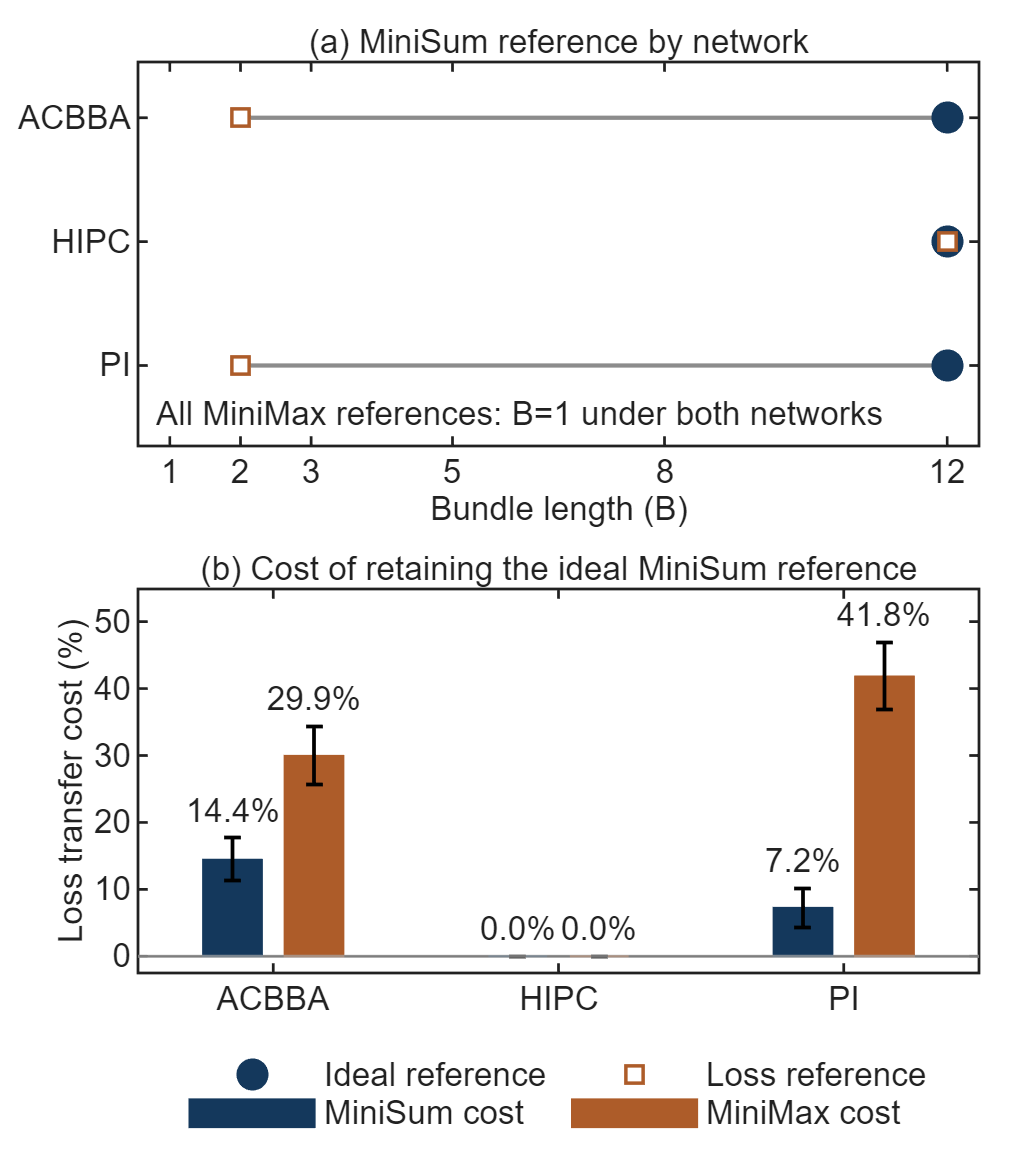}{3.55in}
    \caption{Communication-conditioned MinSum references and their transfer effects. (a) Lowest-mean tested $B$ under ideal delivery and loss. (b) Cost under loss of retaining the ideal MinSum reference instead of the loss-conditioned reference; error bars are paired-bootstrap 95\% intervals. The MinMax consequence evaluates the same two bundle settings. MinMax-specific references remain $B=1$ in both conditions for all three allocators.}
    \label{fig:transfer}
\end{figure}

Consequently, the MinSum reference moves from $B=12$ under ideal communication to $B=2$ under loss for ACBBA and PI (Fig.~\ref{fig:transfer}). Retaining the ideal reference under loss costs 14.4\% MinSum performance for ACBBA (95\% CI 11.2--17.7\%) and 7.2\% for PI (4.4--10.2\%). It simultaneously raises MinMax cost by 30.0\% and 41.8\%, respectively, relative to $B=2$. For these allocators, the nominal MinSum setting is therefore dominated under the tested loss condition rather than merely expressing a different objective preference.

The held-out analysis confirms the two shifts. Training folds select $(B_{\mathrm{ideal}},B_{\mathrm{loss}})=(12,2)$ in 498 of 500 ACBBA folds and all 500 PI folds. Their mean out-of-fold MinSum penalties are 14.4\% and 7.2\%, with MinMax consequences of 30.0\% and 41.8\%. HIPC's full-sample reference remains $B=12$. Its loss surface is nearly flat between $B=8$ and $B=12$ (65.81 versus 65.74 steps), so training-fold selectors alternate between them and produce no held-out benefit from retuning. For MinMax itself, $B=1$ is selected under both network conditions in all 500 folds for every allocator. Configuration shift is therefore objective- and allocator-specific. 

\subsection{Sensitivity to Target Load}

The five- and twenty-target experiments largely preserve the operating regions observed at ten targets. Under ideal communication, all three allocators continue to favor deep bundles for MinSum and shallow bundles for MinMax. The few changes in the exact lowest-mean bundle are negligible: at twenty targets, PI obtains nearly identical MinSum means at $B=8$ and $B=12$ (89.76 and 89.80), while ACBBA's MinMax reference moves from $B=1$ to $B=3$ by only 0.32 steps. Target load therefore does not materially change the ideal-communication objective response.

The communication-dependent MinSum shift also persists across target loads. For ACBBA and PI, packet loss moves the preferred MinSum region from deep to shallow bundles at five, ten, and twenty targets, whereas HIPC remains in the deep-bundle region. MinMax remains stable at five and ten targets. At twenty targets, ACBBA exhibits a meaningful MinMax shift from $B=3$ to $B=2$ (35.28 to 33.48 steps), while PI's corresponding 0.2-step difference is effectively a tie. Thus, the additional loads consistently reproduce the MinSum interaction, with limited evidence that it extends to MinMax at higher task loads.

\section{Discussion}

\subsection{Interpretation of the Bundle-Length Response}

The opposing MinSum and MinMax responses agree with prior multi-robot routing results showing that MinSum allocation can reduce aggregate travel while MinMax allocation reduces last-task completion time \cite{Patil2022}, and with auction formulations that use objective-specific bidding rules for the two criteria \cite{Otte2020}. The present results show how this established objective distinction manifests during configuration of decentralized multi-task allocators. Increasing $B$ permits longer, spatially coherent routes, reducing aggregate movement while concentrating tasks and travel on fewer robots. The accompanying increases in target- and movement-workload Gini support this route-consolidation interpretation. MinSum and MinMax therefore do not merely evaluate the same configured behavior differently; they favor systematically different operating regions.

The response under packet loss depends on how each allocator constructs and reconciles its bundles. ACBBA and PI construct locally owned paths and use exchanged bid or significance information to resolve competing assignments. With deeper bundles, dropped messages can leave more stale commitments unresolved, weakening the route-efficiency benefit that larger $B$ provides under ideal communication. This is consistent with their loss-conditioned MinSum curves, which reach their lowest means at $B=2$ and then rise.

HIPC instead predicts a multi-agent allocation from locally available team information and applies plan consensus to reconcile the resulting assignments. This combination of implicit coordination and consensus was developed specifically for operation with imperfect situational awareness \cite{HIPC}. Because part of the task partition is reconstructed locally, HIPC may be less dependent on every explicit bundle exchange for maintaining a coherent team allocation. Packet loss still increases its absolute effort, but it does not reverse the MinSum benefit of deeper planning. Its loss-conditioned surface becomes nearly flat between $B=8$ and $B=12$, indicating a stable deep operating region rather than a uniquely determined optimum at one endpoint.

\subsection{Implications for Communication-Robust Evaluation}

A fixed-configuration communication experiment and a condition-specific performance comparison answer different questions. Holding $B$ fixed measures the robustness of a nominally configured algorithm and remains appropriate when that configuration is the treatment of interest. It does not, however, establish the best performance attainable by the allocator under the degraded condition. For ACBBA and PI under MinSum, retaining the ideal-network reference produces greater effort and makespan than the loss-conditioned reference; the nominal setting is therefore dominated under the tested channel. For HIPC, and for MinMax in the primary ten-target experiment, the preferred operating region transfers even though absolute performance degrades.

Accordingly, degraded-communication benchmarks should state whether they evaluate the transfer of a nominal configuration or compare algorithms after condition-specific configuration. Claims about nominal robustness require no retuning. Claims about condition-specific best performance should either configure each allocator using separate data from the relevant communication regime or demonstrate that the nominal response transfers. At minimum, studies should report the tested bundle range, the configuration objective, the communication condition used for selection, and enough of the response to distinguish a substantive shift from a near-tie.

\subsection{Limitations}

This study is limited to grid-based simulation with four homogeneous robots, static known targets, three allocators, six bundle lengths, and a single independent packet-loss level. The quantitative responses may therefore differ under burst loss, latency, bandwidth constraints, obstacles, dynamic tasks, heterogeneous teams, or alternative implementations. The 5- and 20-target extensions contain 25 scenarios per condition and provide exploratory evidence of recurrence rather than confirmatory estimates.

\section{Conclusion}

Bundle length governs a consequential tradeoff between aggregate route efficiency and parallel mission completion. Under ideal communication, increasing $B$ from 1 to 12 reduced MinSum cost by 19.0\% for ACBBA, 31.8\% for HIPC, and 23.0\% for PI, while increasing MinMax cost by 45.6\%, 67.6\%, and 94.3\%, respectively. Workload concentration is consistent with longer, consolidated routes as the mechanism linking these responses.

Communication degradation alters this configuration relationship rather than simply increasing cost uniformly. For ACBBA and PI, the ideal-network MinSum reference becomes dominated under 25\% packet loss, with held-out MinSum penalties of 14.4\% and 7.2\% when it is retained. HIPC continues to favor a deep MinSum region, consistent with its use of local team prediction and plan consensus, while its absolute performance still degrades. The ACBBA and PI MinSum shifts recur at both additional target loads, although the affected objectives and practical magnitudes vary. Bundle length should therefore be reported together with its configuration objective and communication condition, and degraded-network evaluations should distinguish robustness of a fixed nominal configuration from condition-specific allocator performance.

\section*{Data and Code Availability}

\begingroup\raggedright
Simulator source, paired scenarios, raw and combined outputs, validation records, and analysis code are available at
\url{https://github.com/jlott22/multi-task-plan-depth-study}. 
\endgroup

\bibliographystyle{IEEEtran}
\bibliography{references}

\end{document}